\documentclass{article}

\usepackage[preprint]{neurips_2019}

\usepackage[utf8]{inputenc} % allow utf-8 input
\usepackage[T1]{fontenc}    % use 8-bit T1 fonts
\usepackage{hyperref}       % hyperlinks
\usepackage{url}            % simple URL typesetting
\usepackage{booktabs}       % professional-quality tables
\usepackage{amsfonts}       % blackboard math symbols
\usepackage{nicefrac}       % compact symbols for 1/2, etc.
\usepackage{microtype}      % microtypography
\usepackage{graphicx}
\usepackage{caption}
\usepackage{subcaption}
\usepackage{placeins}
\setcitestyle{numbers,square}

\title{Compositional generalization in a deep seq2seq model by separating syntax and semantics}

\author{%
  Jake Russin \\
  Department of Psychology and Neuroscience\\
  University of Colorado Boulder\\
  \texttt{jacob.russin@colorado.edu} \\
\And 
  Jason Jo \\
  MILA\\
  Universit\'e de Montr\'eal \\
\And 
  Randall C. O'Reilly \\
  Department of Psychology and Neuroscience\\
  University of Colorado Boulder \\
\And 
  Yoshua Bengio \\
  MILA, Universit\'e de Montr\'eal\\
  CIFAR Senior Fellow \\
}  

\begin{document}

\maketitle

\begin{abstract}
Standard methods in deep learning for natural language processing fail to capture the compositional structure of human language that allows for systematic generalization outside of the training distribution. However, human learners readily generalize in this way, e.g. by applying known grammatical rules to novel words. Inspired by work in neuroscience suggesting separate brain systems for syntactic and semantic processing, we implement a modification to standard approaches in neural machine translation, imposing an analogous separation. The novel model, which we call Syntactic Attention, substantially outperforms standard methods in deep learning on the SCAN dataset, a compositional generalization task, without any hand-engineered features or additional supervision. Our work suggests that separating syntactic from semantic learning may be a useful heuristic for capturing compositional structure. 
\end{abstract}

\section{Introduction}

A crucial property underlying the expressive power of human language is its systematicity  \citep{LakeUllmanTenenbaumEtAl17,FodorPylyshyn88}: syntactic or grammatical rules allow arbitrary elements to be combined in novel ways, making the number of sentences possible in a language to be exponential in the number of its basic elements.
Recent work has shown that standard deep learning methods in natural language processing fail to capture this important property: when tested on unseen combinations of known elements, state-of-the-art models fail to generalize \citep{LakeBaroni17b, LoulaBaroniLake18, BastingsBaroniWestonEtAl18}. 
It has been suggested that this failure represents a major deficiency of current deep learning models, especially when they are compared to human learners \citep{Marcus18,LakeUllmanTenenbaumEtAl17}.

A recently published dataset called SCAN \citep{LakeBaroni17b} (\textbf{S}implified version of the \textbf{C}omm\textbf{A}I \textbf{N}avigation tasks), tests compositional generalization in a sequence-to-sequence (seq2seq) setting by systematically holding out of the training set all inputs containing a basic primitive verb ("jump"), and testing on sequences containing that verb. 
Success on this difficult problem requires models to generalize knowledge gained about the other primitive verbs ("walk", "run" and "look") to the novel verb "jump," without having seen "jump" in any but the most basic context ("jump" $\rightarrow$ JUMP). 
It is trivial for human learners to generalize in this way (e.g. if I tell you that "dax" is a verb, you can generalize its usage to all kinds of constructions, like "dax twice and then dax again", without even knowing what the word means) \citep{LakeBaroni17b}.
However, standard recurrent seq2seq models fail miserably on this task, with the best-reported model (a gated recurrent unit augmented with an attention mechanism) achieving only 12.5\% accuracy on the test set \citep{LakeBaroni17b,BastingsBaroniWestonEtAl18}. Recently, convolutional neural networks (CNN) were shown to perform better on this test, but still only achieved 69.2\% accuracy on the test set.

\begin{figure}
	\begin{center}
	\includegraphics[width=0.6\linewidth]{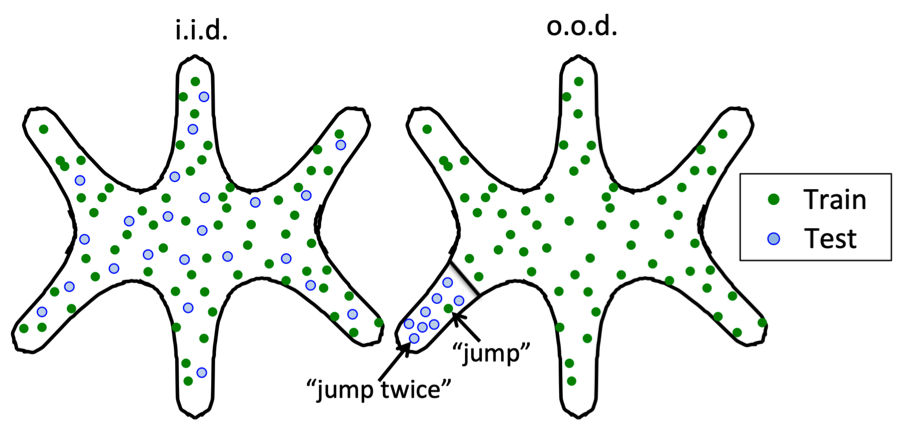}
	\caption{Simplified illustration of out-of-domain (o.o.d.) extrapolation required by SCAN compositional generalization task. Shapes represent the distribution of all possible command sequences. In a simple split, train and test data are independent and identically distributed (i.i.d.), but in the add-primitive splits, models are required to extrapolate out-of-domain from a single example.}
	\label{fig:iid_ood}
	\end{center}
\end{figure}

From a statistical-learning perspective, this failure is quite natural. The neural networks trained on the SCAN task fail to generalize because they have memorized biases that do indeed exist in the training set. Because "jump" has never been seen with any adverb, it would not be irrational to assume that "jump twice" is an invalid sentence in this language. The SCAN task requires networks to make an inferential leap about the entire structure of part of the distribution that they have not seen - that is, it requires them to make an out-of-domain (o.o.d.) \textit{extrapolation} \citep{Marcus18}, rather than merely \textit{interpolate} according to the assumption that train and test data are independent and identically distributed (i.i.d.) (see Figure \ref{fig:iid_ood}). Seen another way, the SCAN task and its analogues in human learning (e.g. "dax"), require models \textit{not} to learn some of the correlations that are actually present in the training data \citep{KrieteNoelleCohenEtAl13}.

Given that humans can perform well on certain kinds of o.o.d. extrapolation tasks, the human brain must be implementing principles that allow humans to generalize systematically, but which are lacking in current deep learning models. 
One prominent idea from neuroscience research on language processing that may offer such a principle is that the brain contains partially separate systems for processing syntax and semantics. In this paper, we motivate such a separation from a machine-learning perspective, and test a simple implementation on the SCAN dataset. Our novel model, which we call Syntactic Attention, encodes syntactic and semantic information in separate streams before producing output sequences. Our experiments show that our novel architecture achieves substantially improved compositional generalization performance over other recurrent networks on the SCAN dataset.

\subsection{Syntax and prefrontal cortex}

Syntax is the aspect of language underlying its systematicity \citep{FodorPylyshyn88}. When given a novel verb like "dax," humans can generalize its usage to many different constructions that they have never seen before, by applying known syntactic or grammatical rules about verbs (e.g. rules about how to conjugate to a different tense or about how adverbs modify verbs). It has long been thought that humans possess specialized cognitive machinery for learning the syntactic or grammatical structure of language \citep{Chomsky57}. A part of the prefrontal cortex called Broca's area, originally thought only to be involved in language production, was later found to be important for comprehending syntactically complex sentences, leading some to conclude that it is important for syntactic processing in general \citep{CaramazzaZurif76a,Thompson-Schill04}. For example, patients with lesions to this area showed poor comprehension on sentences such as "The girl that the boy is chasing is tall". Sentences such as this one require listeners to process syntactic information because semantics is not enough to understand their meanings - e.g. either the boy or the girl could be doing the chasing, and either could be tall. 
 
A more nuanced view situates the functioning of Broca's area within the context of prefrontal cortex in general, noting that it may simply be a part of prefrontal cortex specialized for language \citep{Thompson-Schill04}. The prefrontal cortex is known to be important for cognitive control, or the active maintenance of top-down attentional signals that bias processing in other areas of the brain \citep{MillerCohen01} (see diagram on the right of Figure \ref{fig:architecture}). In this framework, Broca's area can be thought of as a part of prefrontal cortex specialized for language, and responsible for selectively attending to linguistic representations housed in other areas of the brain \citep{Thompson-Schill04}. 

The prefrontal cortex has received much attention from computational neuroscientists \citep{MillerCohen01,OReillyFrank06}, and one model even showed a capacity for compositional generalization \citep{KrieteNoelleCohenEtAl13}. However, these ideas have not been taken up in deep learning research. Here, we emphasize the idea that the brain contains two separate systems for processing syntax and semantics, where the semantic system learns and stores representations of the meanings of words, and the syntactic system, housed in Broca's area of the prefrontal cortex, learns how to selectively attend to these semantic representations according to grammatical rules. 

\section{Syntactic Attention} \label{sec:SyntacticAttention}
\begin{figure}
	\begin{center}
	\begin{subfigure}{0.6\linewidth}
		\centering
		\includegraphics[width=1.0\linewidth]{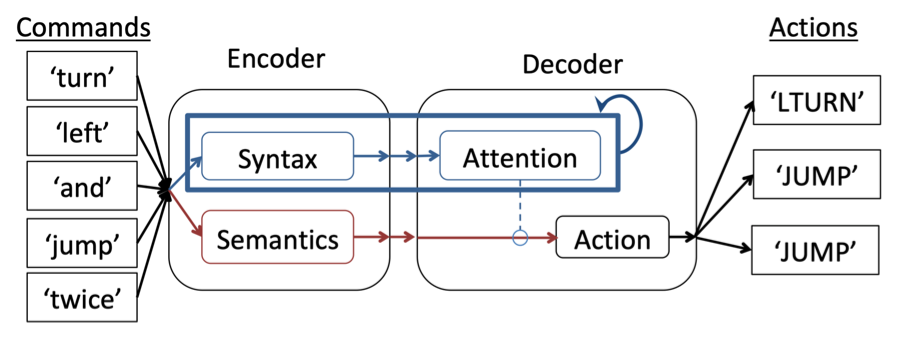}
	\end{subfigure}
	\begin{subfigure}{0.3\linewidth}
		\centering
		\includegraphics[width=1.0\linewidth]{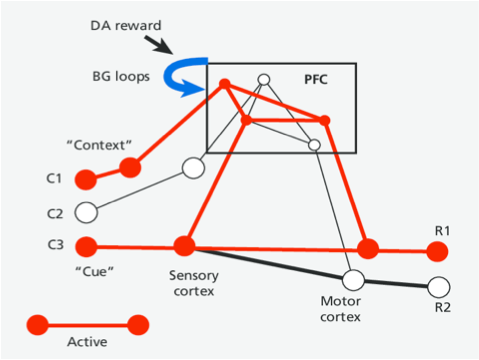}
	\end{subfigure}
	\caption{(left) Syntactic Attention architecture. Syntactic and semantic information are maintained in separate streams. The semantic stream processes words with a simple linear transformation, so that sequential information is not maintained. This information is used to directly produce actions. The syntactic stream processes inputs with a recurrent neural network, allowing it to capture temporal dependencies between words. This stream determines the attention over semantic representations at each time step during decoding. (right) Diagram of an influential computational model of prefrontal cortex (PFC) \citep{MillerCohen01}. Prefrontal cortex dynamically modulates processes in other parts of the brain through top-down selective attention signals. A part of the prefrontal cortex, Broca's area, is thought to be important for syntactic processing \citep{Thompson-Schill04}. Figure reproduced from \citep{Miller13}.}
	\label{fig:architecture}
	\end{center}
\end{figure}

The Syntactic Attention model improves the compositional generalization capability of an existing attention mechanism \citep{BahdanauChoBengio14a} by implementing two separate streams of information processing for syntax and semantics (see Figure \ref{fig:architecture}). Here, by "semantics" we mean the information in each word in the input that determines its \textit{meaning} (in terms of target outputs), and by "syntax" we mean the information contained in the input sequence that should determine the \textit{alignment} of input to target words. We describe the mechanisms of this separation and the other details of the model below, following the notation of \citep{BahdanauChoBengio14a}, where possible.

\subsection{Separation assumption}

In the seq2seq problem, models must learn a mapping from arbitrary-length sequences of inputs $ \mathbf{x} = \{x_1, x_2, ..., x_{T_x}\}$ to arbitrary-length sequences of outputs $ \mathbf{y} = \{y_1, y_2, ..., y_{T_y} \}$: $ p(\mathbf{y} | \mathbf{x}) $. The attention mehcanism of \citep{BahdanauChoBengio14a} models the conditional probability of each target word given the input sequence and previous targets: $p(y_i|y_1, y_2, ..., y_{i-1}, \mathbf{x})$. This is accomplished by processing the input sequence with a recurrent neural network (RNN) in the encoder. The outputs of this RNN are used both for encoding individual words in the input for later translation, and for determining their alignment to targets during decoding. 

The underlying assumption made by the Syntactic Attention architecture is that the dependence of target words on the input sequence can be separated into two independent factors.
One factor, $p(y_i|x_j) $, which we refer to as "semantics," models the conditional distribution from individual words in the input to individual words in the target. Note that, unlike in the model of \citet{BahdanauChoBengio14a}, these $x_j$ do not contain any information about the other words in the input sequence because they are not processed with an RNN. They are 
"semantic" in the sense that they contain the information relevant to translating into the target language. The other factor, $p(j \rightarrow i | \mathbf{x}) $, which we refer to as "syntax," models the conditional probability that word $j$ in the input is relevant to word $i$ in the target sequence, given the entire input sequence. This alignment is accomplished from encodings of the inputs produced by an RNN. This factor is "syntactic" in the sense that it must capture all of the temporal information in the input that is relevant to determining the serial order of outputs. The crucial architectural assumption, then, is that any temporal dependency between individual words in the input that can be captured by an RNN should only be relevant to their alignment to words in the target sequence, and not to the translation of individual words. This assumption will be made clearer in the model description below. 

\subsection{Encoder}

The encoder produces two separate vector representations for each word in the input sequence. Unlike the previous attention model \citep{BahdanauChoBengio14a}), we separately extract the semantic information from each word  with a linear transformation: 
\begin{equation}
\label{eq:lin_semantics}
m_j = W_m x_j,
\end{equation} 
where $W_m$ is a learned weight matrix that multiplies the one-hot encodings $\{x_1, ..., x_{T_x}\}$. 
Note that the semantic representation of each word does not contain any information about the other words in the sentence. 
As in the previous attention mechanism \citep{BahdanauChoBengio14a}, we use a bidirectional RNN (biRNN) to extract what we now interpret as the syntactic information from each word in the input sequence. 
The biRNN produces a vector for each word on the forward pass, $ (\overrightarrow{h_1}, ..., \overrightarrow{h_{T_x})}$, and a vector for each word on the backward pass, $ (\overleftarrow{h_1}, ..., \overleftarrow{h_{T_x})}$. 
The syntactic information (or "annotations" \citep{BahdanauChoBengio14a}) of each word $x_j$ is determined by the two vectors $\overrightarrow{h_{j-1}}$, $\overleftarrow{h_{j+1}}$ corresponding to the words surrounding it: 
\begin{equation}
 h_j = [\overrightarrow{h_{j-1}};\overleftarrow{h_{j+1}}]
\end{equation}

In all experiments, we used a bidirectional Long Short-Term Memory (LSTM) for this purpose. 
Note that because there is no sequence information in the semantic representations, all of the information required to parse (i.e. align) the input sequence correctly (e.g. phrase structure, modifying relationships, etc.) must be encoded by the biRNN.

\subsection{Decoder}
The decoder models the conditional probability of each target word given the input and the previous targets: 
$p(y_i | y_1, y_2, ..., y_{i-1}, \mathbf{x})$, where $y_i$ is the target translation and $\mathbf{x}$ is the whole input sequence. As in the previous model, we use an RNN to determine an attention distribution over the inputs at each time step (i.e. to align words in the input to the current target). However, our decoder diverges from this model in that the mapping from inputs to outputs is performed from a weighted average of the \textit{semantic} representations of the input words:
\begin{equation}
d_i = \sum_{j=1}^{T_x} \alpha_{ij} m_j \qquad p(y_i | y_1, y_2, ..., y_{i-1}, \mathbf{x}) = f(d_i)
\end{equation}

where $f$ is parameterized by a linear function with a softmax nonlinearity, and the $\alpha_{ij}$ are the weights determined by the attention model. We note again that the $m_j$ are produced directly from corresponding $x_j$, and do not depend on the other inputs.
The attention weights are computed by a function measuring how well the syntactic information of a given word in the input sequence aligns with the current hidden state of the decoder RNN, $s_i$:
\begin{equation} 
\alpha_{ij} = \frac{\exp(e_{ij})}{\sum_{k=1}^{T_x}\exp(e_{ik})} \qquad e_{ij} = a(s_{i}, h_j)
\end{equation}

where $e_{ij}$ can be thought of as measuring the importance of a given input word $x_j$ to the current target word $y_i$, and $s_{i}$ is the current hidden state of the decoder RNN.
\citet{BahdanauChoBengio14a} model the function $a$ with a feedforward network, but following \citep{HudsonManning18a}, we choose to use a simple dot product:
\begin{equation}
a(s_{i},h_j) = s_{i} \cdot h_j,
\end{equation}

relying on the end-to-end backpropagation during training to allow the model to learn to make appropriate use of this function. 
Finally, the hidden state of the RNN is updated with the same weighted combination of the \textit{syntactic} representations of the inputs:
\begin{equation}
s_i = g(s_{i-1}, c_{i}) \qquad c_i = \sum_{j=1}^{T_x} \alpha_{ij} h_j
\end{equation}

where $g$ is the decoder RNN, $s_i$ is the current hidden state, and $c_i$ can be thought of as the information in the attended words that can be used to determine what to attend to on the next time step. Again, in all experiments an LSTM was used.

\section{Experiments}

\subsection{SCAN dataset}

\begin{figure}[h]
	\begin{center}
	\includegraphics[width=0.8\linewidth]{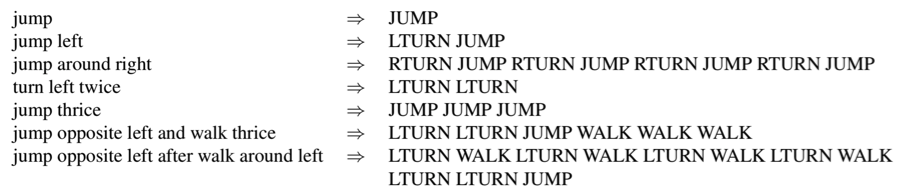}
	\caption{Examples from SCAN dataset. Figure reproduced from \cite{LakeBaroni17b}.}
	\label{fig:SCAN_examples}
	\end{center}
\end{figure}

The SCAN\footnote{The SCAN dataset can be downloaded at \url{https://github.com/brendenlake/SCAN}} dataset is composed of sequences of commands that must be mapped to sequences of actions \citep{LakeBaroni17b} (see Figure \ref{fig:SCAN_examples} and supplementary materials for further details). The dataset is generated from a simple finite phrase-structure grammar that includes things like adverbs and conjunctions. There are 20,910 total examples in the dataset that can be split systematically into training and testing sets in different ways. These splits include the following:
\begin{itemize}
\itemsep0em 
\item Simple split: training and testing data are split randomly
\item Length split: training includes only shorter sequences
\item Add primitive split: a primitive command (e.g. "turn left" or "jump") is held out of the training set, except in its most basic form (e.g. "jump" $\rightarrow$ JUMP) 
\end{itemize}
Here we focus on the most difficult problem in the SCAN dataset, the add-jump split, where "jump" is held out of the training set. The best test accuracy reported in the original paper \citep{LakeBaroni17b}, using standard seq2seq models, was 1.2\%. More recent work has tested other kinds of seq2seq models, including Gated Recurrent Units (GRU) augmented with attention \citep{BastingsBaroniWestonEtAl18} and convolutional neural networks (CNNs) \citep{DessiBaroni19}. Here, we compare the Syntactic Attention model to the best previously reported results.

\subsection{Implementation details}

Experimental procedure is described in detail in the supplementary materials. Train and test sets were kept as they were in the original dataset, but following \citep{BastingsBaroniWestonEtAl18}, we used early stopping by validating on a 20\% held out sample of the training set. All reported results are from runs of 200,000 iterations with a batch size of 1. Unless stated otherwise, each architecture was trained 5 times with different random seeds for initialization, to measure variability in results. All experiments were implemented in PyTorch. Details of the hyperparameter search are given in supplementary materials. Our best model used LSTMs, with 2 layers and 200 hidden units in the encoder, and 1 layer and 400 hidden units in the decoder, and 120-dimensional semantic vectors. The model included a dropout rate of 0.5, and was optimized using an Adam optimizer \citep{KingmaBa14} with a learning rate of 0.001.

\subsection{Results}
The Syntactic Attention model achieves state-of-the-art performance on the key compositional generalization task of the SCAN dataset (see table \ref{tab:table1}). The table shows results (mean test accuracy (\%) $\pm$ standard deviation) on the test splits of the dataset. Syntactic Attention is compared to the previous best models, which were a CNN \citep{DessiBaroni19}, and GRUs augmented with an attention mechanism ("+ attn"), which either included or did not include a dependency ("- dep") in the decoder on the previous action \citep{BastingsBaroniWestonEtAl18}. The best model from the hyperparameter search showed strong compositional generalization performance, attaining a mean accuracy of 91.1\% (median = 98.5\%) on the test set of the add-jump split. However, as in \citet{DessiBaroni19}, we found that our model showed variance across initialization seeds. We suggest that this may be due to the nature of the add-jump split: since "jump" has only been encountered in the simplest context, it may be that slight changes to the way that this verb is encoded can make big differences when models are tested on more complicated constructions. For this reason, we ran the best model 25 times on the add-jump split to get a more accurate assessment of performance. These results were highly skewed, with a mean accuracy of 78.4 \% but a median of 91.0 \% (see supplementary materials for detailed results). Overall, this represents an improvement over the best previously reported results on this task \citep{BastingsBaroniWestonEtAl18,DessiBaroni19}, and does so without any hand-engineered features or additional supervision.  

\begin{table*}[h]
  \begin{center}
  \caption{Compositional generalization results. The Syntactic Attention model achieves an improvement on the compositional generalization tasks of the SCAN dataset, compared to the best previously reported models \citep{BastingsBaroniWestonEtAl18, DessiBaroni19}. Star\textsuperscript{*} indicates median of 25 runs.}
    \begin{tabular}{lcccc}
      Model & \textbf{Simple} & \textbf{Length} & \textbf{Add turn left} & \textbf{Add jump}\\
      \hline
      GRU + attn \citep{BastingsBaroniWestonEtAl18} & 100.0 $\pm$ 0.0 & 18.1 $\pm$ 1.1 & 59.1 $\pm$ 16.8 & 12.5 $\pm$ 6.6 \\
      GRU + attn - dep \citep{BastingsBaroniWestonEtAl18} & 100.0 $\pm$ 0.0 & 17.8 $\pm$ 1.7 & 90.8 $\pm$ 3.6 & 0.7 $\pm$ 0.4\\
      CNN \citep{DessiBaroni19} & 100.0 $\pm$ 0.0 & - & - & 69.2 $\pm$ 8.2 \\
      Syntactic Attention & 100.0 $\pm$ 0.0 & 15.2 $\pm$ 0.7 & \textbf{99.9 $\pm$ 0.16} & \textbf{91.0\textsuperscript{*} $\pm$ 27.4} \\
    \end{tabular}
    \label{tab:table1}
  \end{center}
\end{table*}

\subsection{Additional experiments}
To test our hypothesis that compositional generalization requires a separation between syntax (i.e. sequential information used for alignment), and semantics (i.e. the mapping from individual source words to individual targets), we conducted two more experiments:
\begin{itemize}
\item \textit{Sequential semantics}. An additional biLSTM was used to process the semantics of the sentence: $m_j = [\overrightarrow{m_j};\overleftarrow{m_j}]$, where $\overrightarrow{m_j}$ and $\overleftarrow{m_j}$ are the vectors produced for the source word $x_j$ by a biLSTM on the forward and backward passes, respectively. These $m_j$ replace those generated by the simple linear layer in the Syntactic Attention model (in equation (\ref{eq:lin_semantics})). 
\item  \textit{Syntax-action}. Syntactic information was allowed to directly influence the output at each time step in the decoder: $p(y_i|y_1, y_2, ..., y_{i-1}, \mathbf{x}) = f([d_i; c_i])$, where again $f$ is parameterized with a linear function and a softmax output nonlinearity.
\end{itemize}

The results of the additional experiments (mean test accuracy (\%) $\pm$ standard deviations) are shown in table \ref{tab:table2}. These results partially confirmed our hypothesis: performance on the jump-split test set was worse when the strict separation between syntax and semantics was violated by allowing sequential information to be processed in the semantic stream. However, "syntax-action," which included sequential information produced by a biLSTM (in the syntactic stream) in the final production of actions, maintained good compositional generalization performance. We hypothesize that this was because in this setup, it was easier for the model to learn to use the semantic information to directly translate actions, so it largely ignored the syntactic information. This experiment suggests that the separation between syntax and semantics does not have to be perfectly strict, as long as non-sequential semantic representations are available for direct translation.

\begin{table*}[h]
  \begin{center}
  \caption{Results of additional experiments. Star\textsuperscript{*} indicates median of 25 runs.}
    \begin{tabular}{lcccc}
      Model & \textbf{Simple} & \textbf{Length} & \textbf{Add turn left} & \textbf{Add jump}\\
      \hline
      \textit{Sequential semantics}  & 99.3 $\pm$ 0.7 & 13.1 $\pm$ 2.5 & 99.4 $\pm$ 1.1 & 42.3 $\pm$ 32.7 \\
      \textit{Syntax-action} & 99.3 $\pm$ 0.85 & 15.2 $\pm$ 1.9 & 98.2 $\pm$ 2.2  & 88.7 $\pm$ 14.2 \\
	 Syntactic Attention & 100.0$\pm$ 0.0 & 15.2 $\pm$ 0.7 & 99.9 $\pm$ 0.16 & 91.0\textsuperscript{*} $\pm$ 27.4 \\
    \end{tabular}
    \label{tab:table2}
  \end{center}
\end{table*}

\section{Dicussion}
The Syntactic Attention model was designed to incorporate a key principle that has been hypothesized to describe the organization of the linguistic brain: mechanisms for learning rule-like or syntactic information are separated from mechanisms for learning semantic information. Our experiments confirm that this simple organizational principle encourages systematicity in recurrent neural networks in the seq2seq setting, as shown by the substantial improvement in the model's performance on the compositional generalization tasks in the SCAN dataset. 

The model makes the assumption that the translation of individual words in the input should be independent of their alignment to words in the target sequence. To this end, two separate encodings are produced for the words in the input: semantic representations in which each word is not influenced by other words in the sentence, and syntactic representations which are produced by an RNN that can capture temporal dependencies in the input sequence (e.g. modifying relationships, binding to grammatical roles). Just as Broca's area of the prefrontal cortex is thought to play a role in syntactic processing through a dynamic selective-attention mechanism that biases processing in other areas of the brain, the syntactic system in our model encodes serial information and is constrained to influence outputs through an attention mechanism alone. 

Patients with lesions to Broca's area are able to comprehend sentences like "The girl is kicking a green ball", where semantics can be used to infer the grammatical roles of the words (e.g. that the girl, not the ball, is doing the kicking) \citep{CaramazzaZurif76a}. However, these patients struggle with sentences such as "The girl that the boy is chasing is tall", where the sequential order of the words, rather than semantics, must be used to infer grammatical roles (e.g. either the boy or the girl could be doing the chasing). In our model, the syntactic stream can be seen as analogous to Broca's area, because without it the model would not be able to learn about the temporal dependencies that determine the grammatical roles of words in the input. 

The separation of semantics and syntax, which is in the end a \textit{constraint}, forces the model to learn, in a relatively independent fashion, 1) the individual meanings of words and 2) how the words are being used in a sentence (e.g. how they can modify one another, what grammatical role each is playing, etc.). This encourages systematic generalization because, even if a word has only been encountered in a single context (e.g. "jump" in the add-jump split), as long as its syntactic role is known (e.g. that it is a verb that can be modified by adverbs such as "twice"), it can be used in many other constructions that follow the rules for that syntactic role (see supplementary materials for visualizations). Additional experiments confirmed this intuition, showing that when sequential information is allowed to be processed by the semantic system ("sequential semantics"), systematic generalization performance is substantially reduced. 

\begin{figure}
	\begin{center}
	\includegraphics[width=0.6\linewidth]{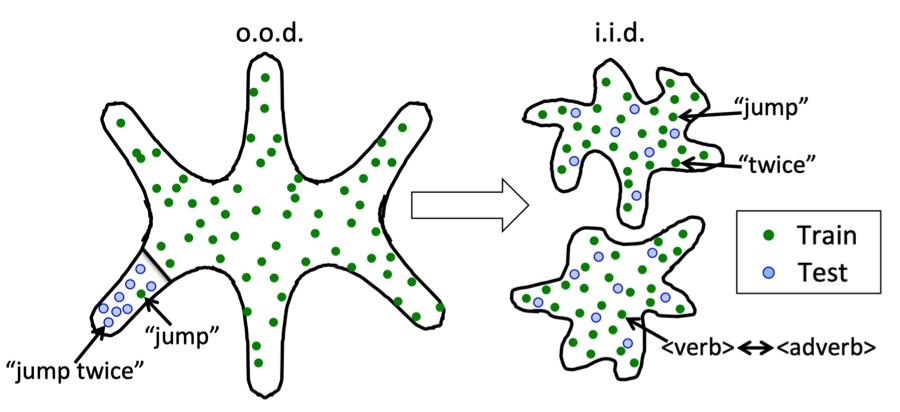}
	\caption{Illustration of the transformation of an out-of-domain (o.o.d.) generalization problem into two independent, identically distributed (i.i.d.) generalization problems. This transformation is accomplished by the Syntactic Attention model without hand-coding grammatical rules or supervising with additional information such as parts-of-speech tags.}
	\label{fig:ood2iid}
	\end{center}
\end{figure}

The Syntactic Attention model bears some resemblance to a symbolic system - the paradigm example of systematicity - in the following sense: in symbolic systems, representational content (e.g. the value of a variable stored in memory) is maintained separately from the computations that are performed on that content.  This separation ensures that the \textit{manipulation} of the content stored in variables is fairly independent of the content itself, and will therefore generalize to arbitrary elements. Our model implements an analogous separation, but in a purely neural architecture that does not rely on hand-coded rules or additional supervision. In this way, it can be seen as transforming a difficult out-of-domain (o.o.d.) generalization problem into two separate i.i.d. generalization problems - one where the individual meanings of words are learned, and one where \textit{how words are used} (e.g. how adverbs modify verbs) is learned (see Figure \ref{fig:ood2iid}). 

It is unlikely that the human brain has such a strict separation between semantic and syntactic processing, and in the end, there must be more of an interaction between the two streams. We expect that the separation between syntax and semantics in the brain is only a relative one, but we have shown here that this kind of separation can be useful for encouraging systematicity and allowing for compositional generalization. 

\section{Other related work}
Our model integrates ideas from computational and cognitive neuroscience \citep{Thompson-Schill04, OReillyFrank06,KrieteNoelleCohenEtAl13,MillerCohen01}, into the neural machine translation framework. Much of the work in neural machine translation uses an encoder-decoder framework, where one RNN is used to encode the source sentence, and then a decoder neural network decodes the representations given by the encoder to produce the words in the target sentence \citep{SutskeverVinyalsLe14}. Earlier work attempted to encode the source sentence into a single fixed-length vector (the final hidden state of the encoder RNN), but it was subsequently shown that better performance could be achieved by encoding each word in the source, and using an attention mechanism to \textit{align} these encodings with each target word during the decoding process \citep{BahdanauChoBengio14a}. The current work builds directly on this attention model, while incorporating a separation between syntactic and semantic information streams.  

The principle of compositionality has recently regained the attention of deep learning researchers \citep{AndreasRohrbachDarrellEtAl15a,BahdanauMurtyNoukhovitchEtAl18, LakeUllmanTenenbaumEtAl17, LakeBaroni17b,BattagliaHamrickBapstEtAl18b,JohnsonHariharanvanderMaatenEtAl16a} . In particular, the issue has been explored in the visual-question answering (VQA) setting \citep{AndreasRohrbachDarrellEtAl15a,HudsonManning18a,JohnsonHariharanvanderMaatenEtAl16a,PerezStrubdeVriesEtAl17,HuAndreasRohrbachEtAl17a,SantoroRaposoBarrettEtAl17a,YiWuGanEtAl18}.
Many of the successful models in this setting learn hand-coded operations \citep{AndreasRohrbachDarrellEtAl15a,HuAndreasRohrbachEtAl17a}, use highly specialized components \citep{HudsonManning18a,SantoroRaposoBarrettEtAl17a}, or use additional supervision \citep{HuAndreasRohrbachEtAl17a,YiWuGanEtAl18}. In contrast, our model uses standard recurrent networks and simply imposes the additional constraint that syntactic and semantic information are processed in separate streams. 

Some of the recent research on compositionality in machine learning has had a special focus on the use of attention. For example, in the Compositional Attention Network, built for VQA, a strict separation is maintained between the representations used to encode images and the representations used to encode questions  \citep{HudsonManning18a}. This separation is enforced by restricting them to interact only through attention distributions. Our model utilizes a similar restriction, reinforcing the idea that compositionality is enhanced when information from different modalities (in our case syntax and semantics) are only allowed to interact through discrete probability distributions.

Previous research on compositionality in machine learning has also focused on the incorporation of symbol-like processing  into deep learning models \citep{AndreasRohrbachDarrellEtAl15a, HuAndreasRohrbachEtAl17a, YiWuGanEtAl18}. These methods generally rely on hand-coding or additional supervision for the symbolic representations or algorithmic processes to emerge. For example, in neural module networks \citep{AndreasRohrbachDarrellEtAl15a}, a neural network is constructed out of composable neural modules that each learn a specific operation. These networks have shown an impressive capacity for systematic generalization on VQA tasks \citep{BahdanauMurtyNoukhovitchEtAl18}. These models can be seen as accomplishing a similar transformation as depicted in Figure \ref{fig:ood2iid}, because the learning in each module is somewhat independent of the mechanism that composes them. However, \citet{BahdanauMurtyNoukhovitchEtAl18} find that when these networks are trained end-to-end (i.e. without hand-coded parameterizations and layouts) their systematicity is significantly degraded.

In contrast, our model learns in an end-to-end way to generalize systematically without any explicit symbolic processes built in. This offers an alternative way in which symbol-like processing can be achieved with neural networks - by enforcing a separation between mechanisms for learning representational content (semantics) and mechanisms for learning how to dynamically attend to or manipulate that content (syntax) in the context of a cognitive operation or reasoning problem.

\section{Conclusion}

The Syntactic Attention model incorporates ideas from cognitive and computational neuroscience into the neural machine translation framework, and produces the kind of systematic generalization thought to be a key component of human language-learning and intelligence. The key feature of the architecture is the separation of sequential information used for alignment (syntax) from information used for mapping individual inputs to outputs (semantics). This separation allows the model to generalize the usage of a word with known syntax to many of its valid grammatical constructions. This principle may be a useful heuristic in other natural language processing tasks, and in other systematic or compositional generalization tasks. The success of our approach suggests a conceptual link between dynamic selective-attention mechanisms in the prefrontal cortex and the systematicity of human cognition, and points to the untapped potential of incorporating ideas from cognitive science and neuroscience into modern approaches in deep learning and artificial intelligence \citep{MarblestoneWayneKording16a}.

%\subsubsection*{Acknowledgments}

%Use unnumbered third level headings for the acknowledgments. All acknowledgments go at the end of the paper. Do not include acknowledgments in the anonymized submission, only in the final paper.

%\section*{References}
\bibliographystyle{abbrvnat}
\bibliography{SyntacticAttentionNeurIPS}

\pagebreak
\section{Supplementary materials}

\subsection{SCAN dataset details}
\begin{figure}[h]
	\begin{center}
	\includegraphics[width=0.7\linewidth]{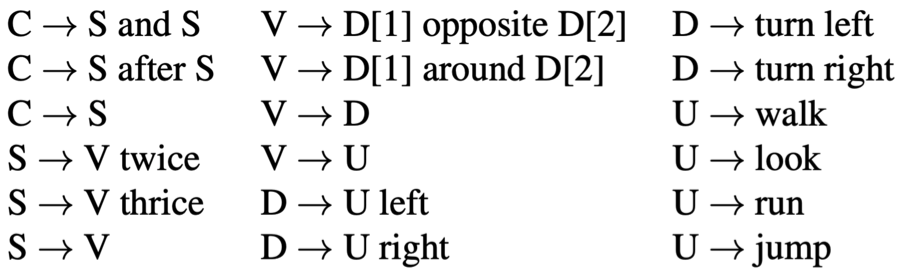}
	\caption{Phrase-structure grammar used to generate SCAN dataset. Figure reproduced from \citep{LakeBaroni17b}.}
	\label{fig:phrase_structure}
	\end{center}
\end{figure}

The SCAN dataset \citep{LakeBaroni17b} generates sequences of commands using the pharase-structure grammar described in Figure \ref{fig:phrase_structure}. This simple grammar is not recursive, and so can generate a finite number of command sequences (20,910 total). 

\begin{figure}[h]
	\begin{center}
	\includegraphics[width=0.9\linewidth]{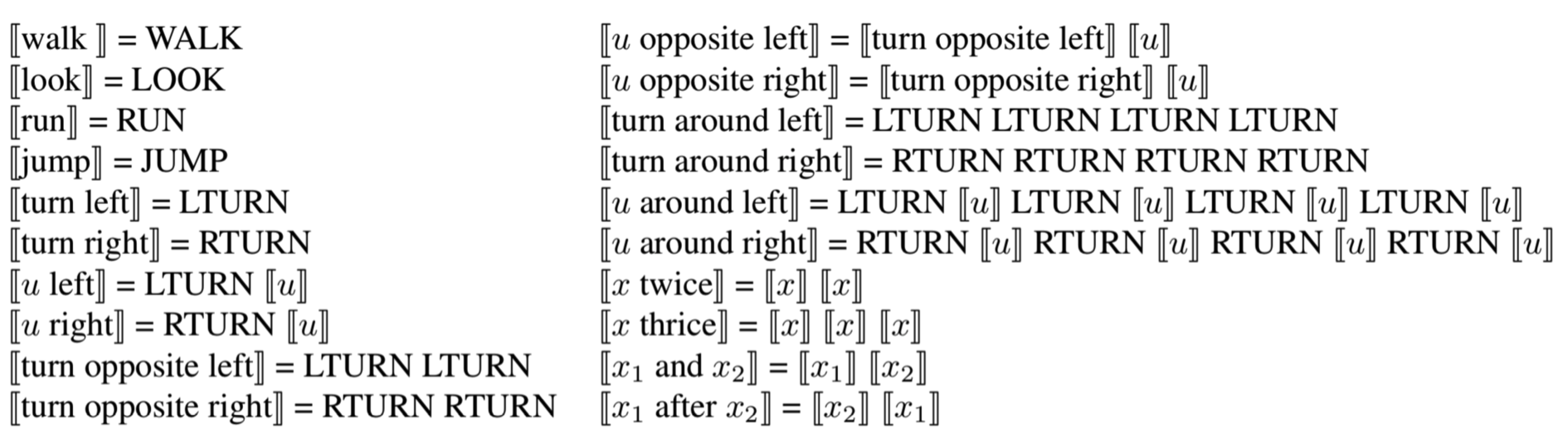}
	\caption{Rules for interpreting command sequences to generate actions in SCAN dataset. Figure reproduced from \citep{LakeBaroni17b}.}
	\label{fig:interpreter}
	\end{center}
\end{figure}

These commands are interpreted according to the rules shown in Figure \ref{fig:interpreter}. Although the grammar used to generate and interpret the commands is simple compared to any natural language, it captures the basic properties that are important for testing compositionality (e.g. modifying relationships, discrete grammatical roles, etc.). The add-primitive splits (described in main text) are meant to be analogous to the capacity of humans to generalize the usage of a novel verb (e.g. "dax") to many constructions \citep{LakeBaroni17b}. 

\subsection{Experimental procedure details}

The cluster used for all experiments consists of 3 nodes, with 68 cores in total (48 times Intel(R) Xeon(R) CPU E5-2650 v4 at
2.20GHz, 20 times Intel(R) Xeon(R) CPU E5-2650 v3 at 2.30GHz), with 128GB of ram each, connected through a 56Gbit infiniband network. It has 8 pascal Titan X GPUs and runs Ubuntu 16.04.

All experiments were conducted with the SCAN dataset as it was originally published \citep{LakeBaroni17b}. No data were excluded, and no preprocessing was done except to encode words in the input and action sequences into one-hot vectors, and to add special tokens for start-of-sequence and end-of-sequence tokens. Train and test sets were kept as they were in the original dataset, but following \citep{BastingsBaroniWestonEtAl18}, we used early stopping by validating on a 20\% held out sample of the training set. All reported results are from runs of 200,000 iterations with a batch size of 1. Except for the additional batch of 25 runs for the add-jump split, each architecture was trained 5 times with different random seeds for initialization, to measure variability in results. All experiments were implemented in PyTorch.

Initial experimentation included different implementations of the assumption that syntactic information be separated from semantic information. After the architecture described in the main text showed promising results, a hyperparameter search was conducted to determine optimization (stochastic gradient descent vs. Adam), RNN-type (GRU vs. LSTM), regularizers (dropout, weight decay), and number of layers (1 vs. 2 layers for encoder and decoder RNNs). We found that the Adam optimizer \citep{KingmaBa14} with a learning rate of 0.001, two layers in the encoder RNN and 1 layer in the decoder RNN, and dropout worked the best, so all further experiments used these specifications. Then, a grid-search was conducted to find the number of hidden units (in both semantic and syntactic streams) and dropout rate. We tried hidden dimensions ranging from 50 to 400, and dropout rates ranging from 0.0 to 0.5. 

The best model used an LSTM with 2 layers and 200 hidden units in the encoder, and an LSTM with 1 layer and 400 hidden units in the decoder, and used 120-dimensional semantic vectors, and a dropout rate of 0.5. The results for this model are reported in the main text. All additional experiments were done with models derived from this one, with the same hyperparameter settings. 

All evaluation runs are reported in the main text: for each evaluation except for the add-jump split, models were trained 5 times with different random seeds, and performance was measured with means and standard deviations of accuracy. For the add-jump split, we included 25 runs to get a more accurate assessment of performance. This revealed a strong skew in the distribution of results, so we included the median as the main measure of performance. Occasionally, the model did not train at all due to an unknown error (possibly very poor random initialization, high learning rate or numerical error). For this reason, we excluded runs in which training accuracy did not get above 10\%. No other runs were excluded.

\subsection{Skew of add-jump results}
\begin{figure}[h]
	\begin{center}
	\includegraphics[width=0.7\linewidth]{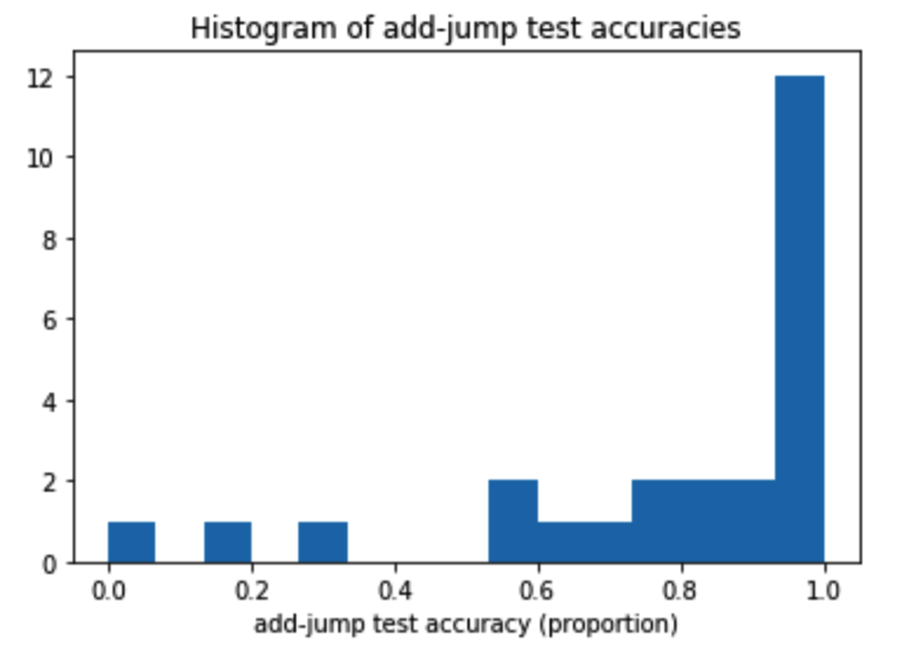}
	\caption{Histogram of test accuracies across all 25 runs of add-jump split.}
	\label{fig:runs_hist}
	\end{center}
\end{figure}

As mentioned in the results section of the main text, we found that test accuracy on the add-jump split was variable and highly skewed. Figure \ref{fig:runs_hist} shows a histogram of these results (proportion correct). The model performs near-perfectly most of the time, but is also prone to catastrophic failures. This may be because, at least for our model, the add-jump split represents a highly nonlinear problem in the sense that slight differences in the way the primitive verb "jump" is encoded during training can have huge differences for how the model performs on more complicated constructions. We recommend that future experiments with this kind of compositional generalization problem take note of this phenomenon, and conduct especially comprehensive analyses of variability in results. Future research will also be needed to better understand the factors that determine this variability, and whether it can be overcome with other priors or regularization techniques.

\subsection{Supplementary experiments}

\subsubsection{Testing nonlinear semantics}
Our main hypothesis is that the separation between sequential information used for alignment (syntax) and information about the meanings of individual words (semantics) encourages systematicity. The results reported in the main text are largely consistent with this hypothesis, as shown by the performance of the Syntactic Attention model on the composotional generalization tests of the SCAN dataset. However, it is also possible that the simplicity of the semantic stream in the model is also important for improving compositional generalization. To test this, we replaced the linear layer in the semantic stream with a nonlinear neural network. From the model description in the main text:
\begin{equation}
p(y_i|y_1, y_2, ..., y_{i-1}, \mathbf{x}) = f(d_i),
\end{equation}
In the original model, $f$ was parameterized with a simple linear layer, but here we use a two-layer feedforward network with a ReLU nonlinearity, before a softmax is applied to generate a distribution over the possible actions. We tested this model on the add-primitive splits of the SCAN dataset. The results (mean (\%) with standard deviations) are shown in Table \ref{tab:sem_mlp}, with comparison to the baseline Syntactic Attention model. 

\begin{table*}[h]
  \begin{center}
  \caption{Results of nonlinear semantics experiment. Star\textsuperscript{*} indicates median of 25 runs.}
    \begin{tabular}{lcc}
      Model  & \textbf{Add turn left} & \textbf{Add jump}\\
      \hline
      \textit{Nonlinear semantics}   & 99.0 $\pm$ 1.7 & 84.4 $\pm$ 14.1 \\
	 Syntactic Attention & 99.9 $\pm$ 0.16 & 91.0\textsuperscript{*} $\pm$ 27.4  \\
    \end{tabular}
    \label{tab:sem_mlp}
  \end{center}
\end{table*}

The results show that this modification did not substantially degrade compositional generalization performance, suggesting that the success of the Syntactic Attention model does not depend on the parameterization of the semantic stream with a simple linear function. 

\subsubsection{Add-jump split with additional examples}
The original SCAN dataset was published with compositional generalization splits that have more than one example of the held-out primitive verb \citep{LakeBaroni17b}. The training sets in these splits of the dataset include 1, 2, 4, 8, 16, or 32 random samples of command sequences with the "jump" command, allowing for a more fine-grained measurement of the ability to generalize the usage of a primitive verb from few examples. For each number of "jump" commands included in the training set, five different random samples were taken to capture any variance in results due to the selection of particular commands to train on. 

\citet{LakeBaroni17b} found that their best model (an LSTM without an attention mechansim) did not generalize well (below 39\%), even when it was trained on 8 random examples that included the "jump" command, but that the addition of further examples to the training set improved performance. Subsequent work showed better performance at lower numbers of "jump" examples, with GRU's augmented with an attention mechanism ("+ attn"), and either with or without a dependence in the decoder on the previous target ("- dep") \citep{BastingsBaroniWestonEtAl18}. Here, we compare the Syntactic Attention model to these results.

\begin{figure}[h]
	\begin{center}
	\includegraphics[width=0.7\linewidth]{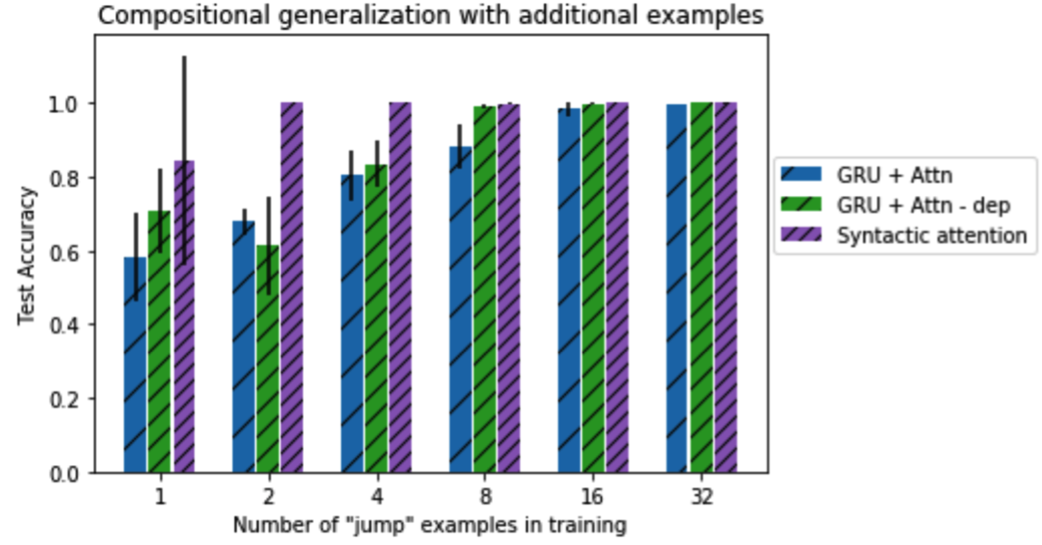}
	\caption{Compositional generalization performance on add-jump split with additional examples. Syntactic Attention model is compared to previously reported models \citep{BastingsBaroniWestonEtAl18} on test accuracy as command sequences with "jump" are added to the training set. Mean accuracy (proportion correct) was computed with 5 different random samples of "jump" commands. Error bars represent standard deviations.}
	\label{fig:WAE}
	\end{center}
\end{figure}

The Syntactic Attention model shows a substantial improvement over previously reported results at the lowest numbers of "jump" examples used for training (see Figure \ref{fig:WAE} and Table \ref{tab:WAE}). Compositional generalization performance is already quite high at 1 example, and at 2 examples is almost perfect (99.997\% correct).

\begin{table*}[h]
  \begin{center}
  \caption{Results of Syntactic Attention compared to models of \citet{BastingsBaroniWestonEtAl18} on jump-split with additional examples. Mean accuracy (\% - rounded to tenths) is shown with standard deviations. Same data as depicted in Figure \ref{fig:WAE}.}
    \begin{tabular}{c cccccc}
	\toprule
	  & \multicolumn{6}{c}{Number of jump commands in training set}\\
	  \cmidrule(r){2-7}
      Model & 1 & 2 & 4 & 8 & 16 & 32 \\
      \midrule
      GRU + attn & 58.2{\tiny$\pm$12.0} & 67.8{\tiny$\pm$3.4} & 80.3{\tiny$\pm$7.0} & 88.0{\tiny$\pm$6.0} & 98.3{\tiny$\pm$1.8} & 99.6{\tiny$\pm$0.2} \\
      GRU + attn - dep & 70.9{\tiny$\pm$11.5} & 61.3{\tiny$\pm$13.5} & 83.5{\tiny$\pm$6.1} & 99.0{\tiny$\pm$0.4} & 99.7{\tiny$\pm$0.2} & 100.0{\tiny$\pm$0.0} \\
      Syntactic Attention & 84.4{\tiny$\pm$28.5} & 100.0{\tiny$\pm$0.01} & 100.0{\tiny$\pm$0.02} & 99.9{\tiny$\pm$0.2} & 100.0{\tiny$\pm$0.01} & 99.9{\tiny$\pm$0.2} \\
    \bottomrule
    \end{tabular}
    \label{tab:WAE}
  \end{center}
\end{table*}

\subsubsection{Template splits}
\begin{figure}[h]
	\begin{center}
	\includegraphics[width=0.9\linewidth]{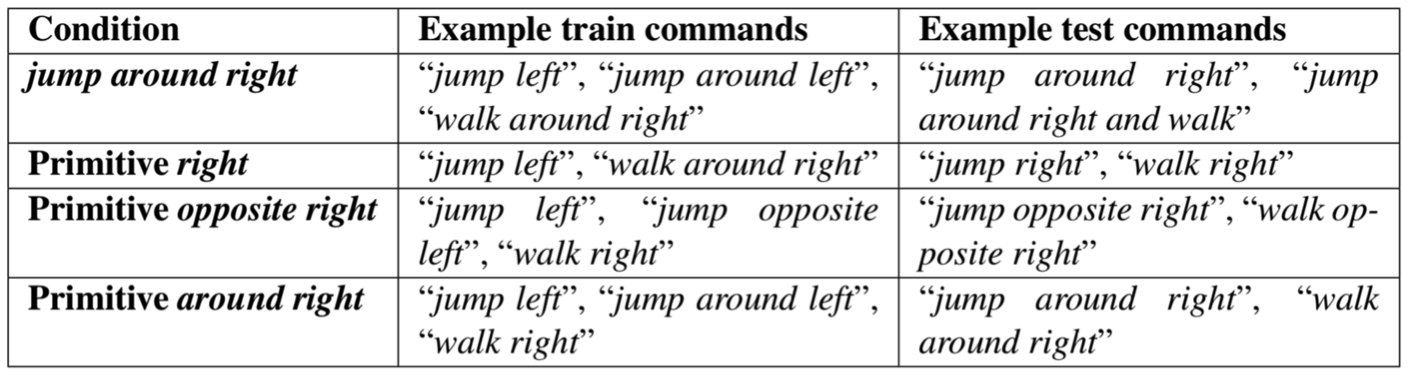}
	\caption{Table of example command sequences for each template split. Reproduced from \citep{LoulaBaroniLake18}}.
	\label{fig:template_examples}
	\end{center}
\end{figure}

The compositional generalization splits of the SCAN dataset were originally designed to test for the ability to generalize known primitive verbs to valid unseen constructions \citep{LakeBaroni17b}. Further work with SCAN augmented this set of tests to include compositional generalization based not on known verbs but on known \textit{templates} \citep{LoulaBaroniLake18}. These template splits included the following (see Figure \ref{fig:template_examples} for examples):
\begin{itemize}
\itemsep0em 
\item \textit{Jump around right}: All command sequences with the phrase "jump around right" are held out of the training set and subsequently tested.
\item Primitive \textit{right}: All command sequences containing primitive verbs modified by "right" are held out of the training set and subsequently tested.
\item Primitive \textit{opposite right}: All command sequences containing primitive verbs modified by "opposite right" are held out of the training set and subsequently tested.
\item Primitive \textit{around right}: All command sequences containing primitive verbs modified by "around right" are held out of the training set and subsequently tested. 
\end{itemize}

Results of the Syntactic Attention model on these template splits are compared to those originally published \citep{LoulaBaroniLake18} in Table \ref{tab:template}. The model, like the one reported in \citep{LoulaBaroniLake18}, performs well on the \textit{jump around right} split, consistent with the idea that this task does not present a problem for neural networks. The rest of the results are mixed: Syntactic Attention shows good compositional generalization performance on the Primitive \textit{right} split, but fails on the Primitive \textit{opposite right} and Primitive \textit{around right} splits. All of the template tasks require models to generalize based on the symmetry between "left" and "right" in the dataset. However, in the \textit{opposite right} and \textit{around right} splits, this symmetry is substantially violated, as one of the two prepositional phrases in which they can occur is never seen with "right." Further research is required to determine whether a model implementing similar principles to Syntactic Attention can perform well on this task.

\begin{table*}[h]
  \begin{center}
  \caption{Results of Syntactic Attention compared to models of \citet{LoulaBaroniLake18} on template splits of SCAN dataset. Mean accuracy (\%) is shown with standard deviations. \textbf{P} = Primitive}
    \begin{tabular}{c cccc}
	\toprule
	  & \multicolumn{4}{c}{Template split}\\
	  \cmidrule(r){2-5}
      Model & \textit{jump around right} & \textbf{P} \textit{right} & \textbf{P} \textit{opposite right} & \textbf{P} \textit{around right} \\
      \midrule
      LSTM (\citet{LoulaBaroniLake18}) & 98.43{$\pm$0.54} & 23.49{$\pm$8.09} & 47.62{$\pm$17.72} & 2.46{$\pm$2.68} \\
      Syntactic Attention & 98.9{$\pm$2.3} & 99.1{$\pm$1.8} & 10.5{$\pm$8.8} & 28.9{$\pm$34.8} \\
    \bottomrule
    \end{tabular}
    \label{tab:template}
  \end{center}
\end{table*}

\subsection{Visualizing attention}
The way that the attention mechanism of \citet{BahdanauChoBengio14a} is set up allows for easy visualization of the model's attention. Here, we visualize the attention distributions over the words in the command sequence at each step during the decoding process. In the following figures (Figures \ref{fig:attn1} to \ref{fig:attn6}), the attention weights on each command (in the columns of the image) is shown for each of the model's outputs (in the rows of the image) for some illustrative examples. Darker blue indicates a higher weight. The examples are shown in pairs for a model trained and tested on the add-jump split, with one example drawn from the training set and a corresponding example drawn from the test set. Examples are shown in increasing complexity, with a failure mode depicted in Figure \ref{fig:attn6}.

In general, it can be seen that although the attention distributions on the test examples are not exactly the same as those from the corresponding training examples, they are usually good enough for the model to produce the correct action sequence. This shows the model's ability to apply the same syntactic rules it learned on the other verbs to the novel verb "jump." In the example shown in Figure \ref{fig:attn6}, the model fails to attend to the correct sequence of commands, resulting in an error. 

\FloatBarrier

\begin{figure}[h]
	\begin{center}
	\begin{subfigure}{0.49\linewidth}
		\centering
		\includegraphics[width=1.0\linewidth]{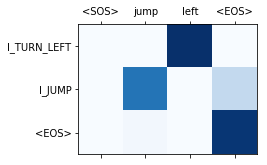}
	\end{subfigure}
	\begin{subfigure}{0.49\linewidth}
		\centering
		\includegraphics[width=1.0\linewidth]{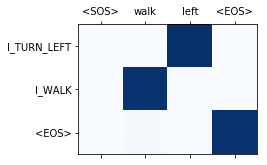}
	\end{subfigure}
	\caption{Attention distributions: correct example}
	\label{fig:attn1}
	\end{center}
\end{figure}

\begin{figure}[h]
	\begin{center}
	\begin{subfigure}{0.49\linewidth}
		\centering
		\includegraphics[width=1.0\linewidth]{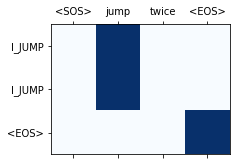}
	\end{subfigure}
	\begin{subfigure}{0.49\linewidth}
		\centering
		\includegraphics[width=1.0\linewidth]{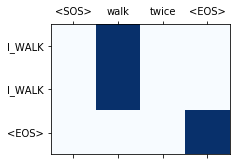}
	\end{subfigure}
	\caption{Attention distributions: correct example}
	\label{fig:attn2}
	\end{center}
\end{figure}

\begin{figure}[h]
	\begin{center}
	\begin{subfigure}{0.49\linewidth}
		\centering
		\includegraphics[width=1.0\linewidth]{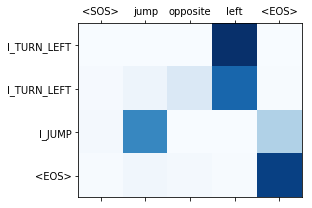}
	\end{subfigure}
	\begin{subfigure}{0.49\linewidth}
		\centering
		\includegraphics[width=1.0\linewidth]{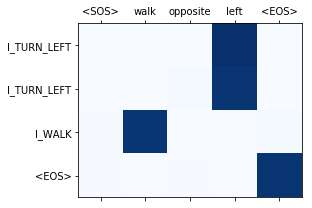}
	\end{subfigure}
	\caption{Attention distributions: correct example}
	\label{fig:attn3}
	\end{center}
\end{figure}

\begin{figure}[t!]
	\begin{center}
	\begin{subfigure}{0.4\linewidth}
		\centering
		\includegraphics[width=1.0\linewidth]{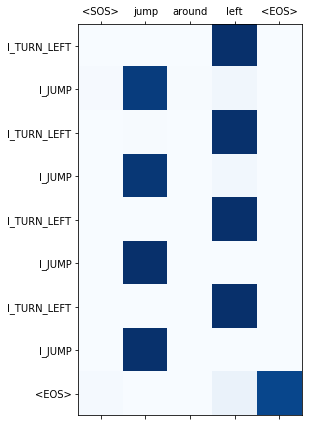}
	\end{subfigure}
	\begin{subfigure}{0.4\linewidth}
		\centering
		\includegraphics[width=1.0\linewidth]{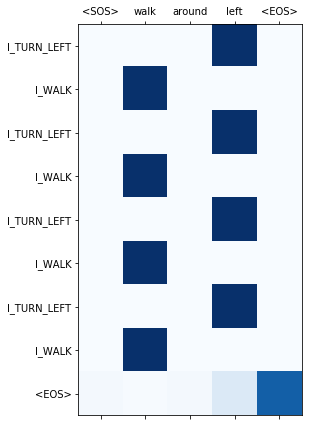}
	\end{subfigure}
	\caption{Attention distributions: correct example}
	\label{fig:attn4}
	\end{center}
\end{figure}

\begin{figure}[b!]
	\begin{center}
	\begin{subfigure}{0.49\linewidth}
		\centering
		\includegraphics[width=1.0\linewidth]{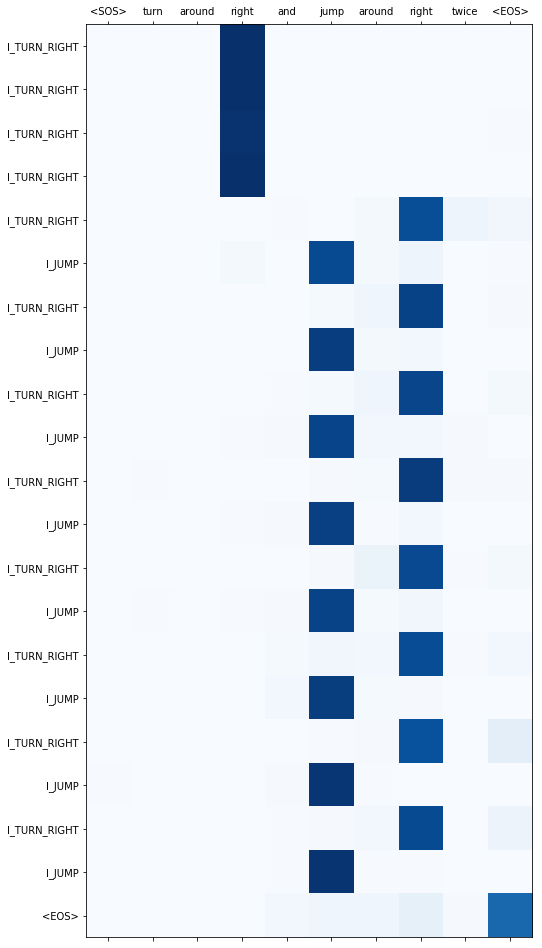}
	\end{subfigure}
	\begin{subfigure}{0.49\linewidth}
		\centering
		\includegraphics[width=1.0\linewidth]{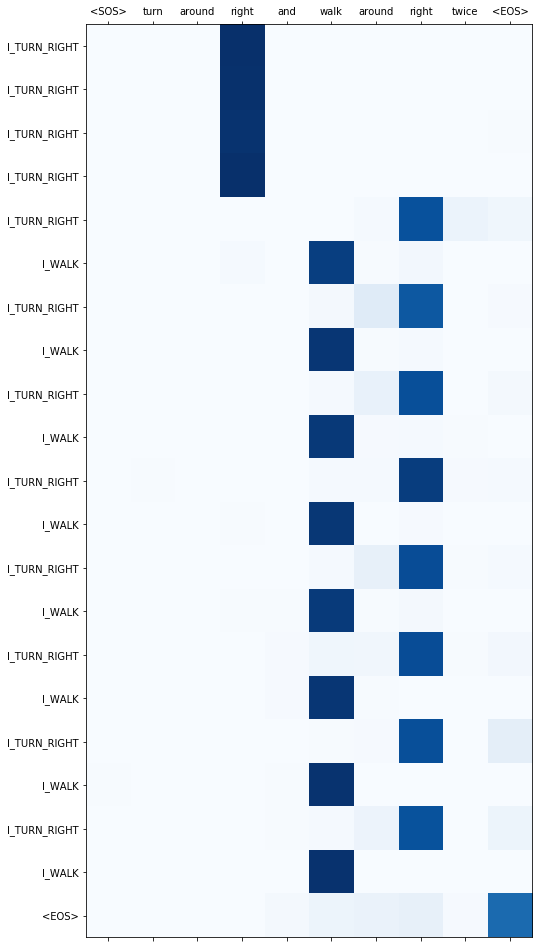}
	\end{subfigure}
	\caption{Attention distributions: correct example}
	\label{fig:attn5}
	\end{center}
\end{figure}

\begin{figure}[t]
	\begin{center}
	\begin{subfigure}{0.9\linewidth}
		\centering
		\includegraphics[width=1.0\linewidth]{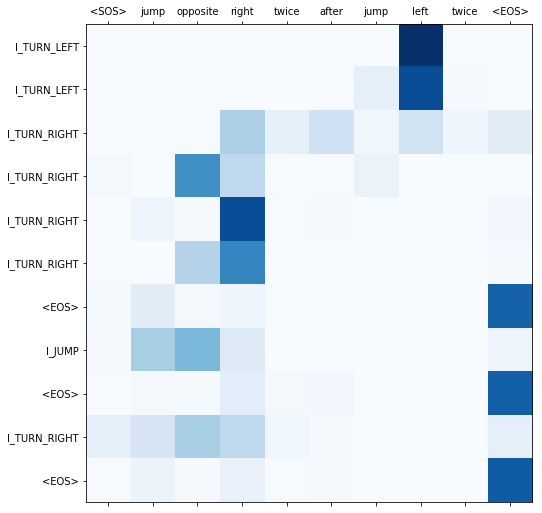}
	\end{subfigure}
	\begin{subfigure}{0.9\linewidth}
		\centering
		\includegraphics[width=1.0\linewidth]{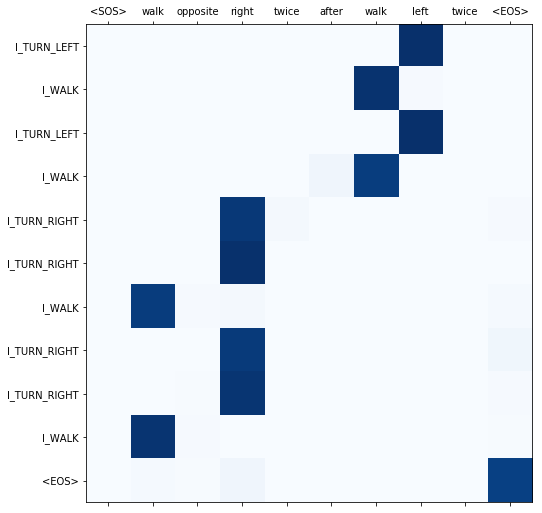}
	\end{subfigure}
	\caption{Attention distributions: incorrect example}
	\label{fig:attn6}
	\end{center}
\end{figure}

\end{document}